\documentclass{article}

\usepackage[preprint]{corl_2024} 

\title{MetaCropFollow: Few-Shot Adaptation with Meta-Learning for Under-Canopy Navigation}

\usepackage{authblk}

\author[1, 2]{Thomas Woehrle}
\author[2]{Arun N. Sivakumar}
\author[2]{Naveen Uppalapati}
\author[2]{Girish Chowdhary}

\affil[1]{Department of Computer Science, Technical University of Munich}
\affil[2]{Field Robotics Engineering and Sciences Hub (FRESH), University of Illinois Urbana-Champaign}

\usepackage{amsmath}
\usepackage{array}
\usepackage{bm}
\usepackage{booktabs}
\usepackage{graphicx}

\begin{document}

\setlength{\abovecaptionskip}{10pt} 

\maketitle

\begin{figure}[h]
    \centering
    \includegraphics[width=0.5\textwidth]{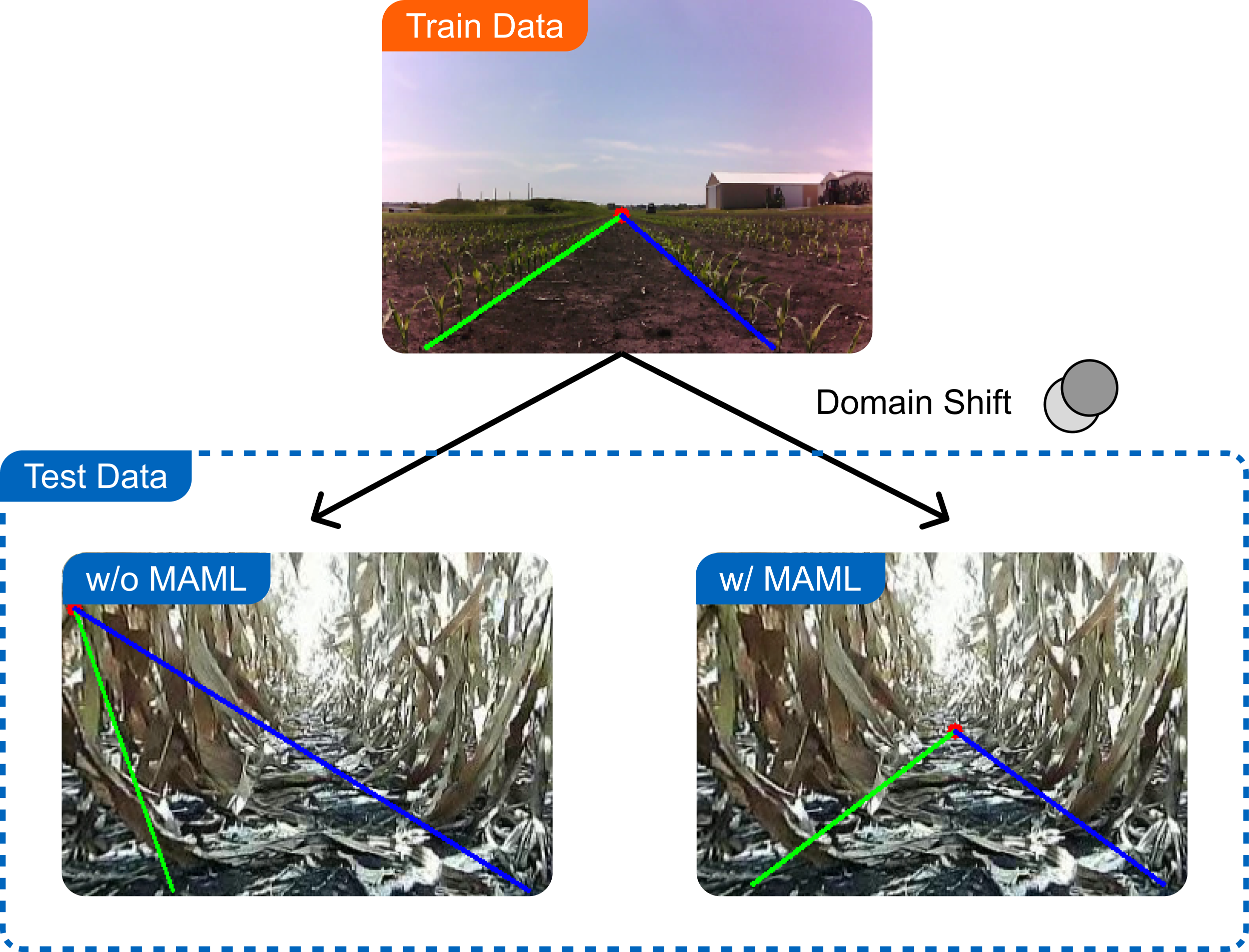}
    \caption{Visual under-canopy agricultural navigation is challenging due to large variations in lighting, appearance of the crops and soil throughout the season. We evaluate the use of MAML for few-shot adaptation of a semantic keypoint prediction network introduced by \citep{Sivakumar-RSS-24}.}
    \label{fig:top_figure}
\end{figure}

\begin{abstract}
    Autonomous under-canopy navigation faces additional challenges compared to over-canopy settings - for example the tight spacing between the crop rows, degraded GPS accuracy and excessive clutter. Keypoint-based visual navigation has been shown to perform well in these conditions, however the differences between agricultural environments in terms of lighting, season, soil and crop type mean that a domain shift will likely be encountered at some point of the robot deployment. In this paper, we explore the use of Meta-Learning to overcome this domain shift using a minimal amount of data. We train a base-learner that can quickly adapt to new conditions, enabling more robust navigation in low-data regimes. 
\end{abstract}

\keywords{Agricultural Robotics, Meta-Learning, Vision-based Navigation} 


\section{Introduction}
    Autonomous agricultural robots promise to help solve the problems of higher food production demands, reduced availability of farm labour and higher needs for agricultural sustainability \citep{RaminShamshiri2018, Foley2011}. Broadly speaking, such robots operate either in over-canopy or under-canopy settings, with the latter being less common but promising, particularly in the case of plant-level monitoring and care. However, under-canopy conditions bring with them certain challenges, which robots operating over-canopy do not face. Over-canopy robots (e.g. drones, tractors, and combine harvesters) oftentimes make use of RTK (Real-Time Kinematic)-GPS, which can not be reliably used under-canopy because of signal attenuation and multi-path error \citep{Sivakumar2021, farrel2008}. Tight-spaced rows and regular occlusions because of densely growing plants constitute additional difficulties \citep{Sivakumar2021}. 
    \par
    Visual navigation using cameras can provide a robust and low-cost solution to these problems. One such solution is CropFollow++ introduced by \citet{Sivakumar-RSS-24}, which learns to identify keypoints to represent the navigation problem. Namely, these key-points are the vanishing point, left intersect-point with the x-axis and right intersect-point with the x-axis. Combined they create a triangle representing the traversable area between two rows of crop. 
    \par
\begin{figure}[t]
    \centering
    \includegraphics[width=\textwidth]{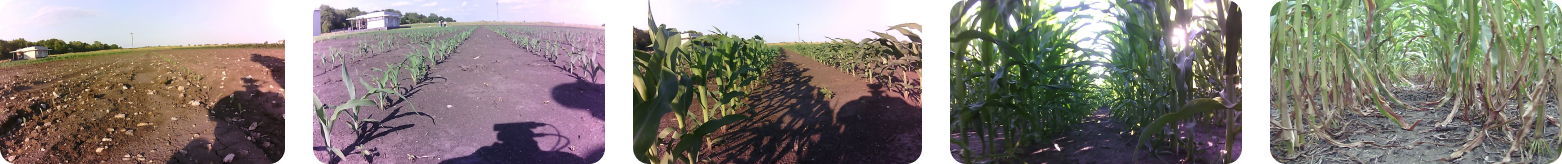}
    \caption{Samples of the dataset used.}
    \label{fig:data_samples}
\end{figure}

 
    However, the variation across different fields, plant types and seasons (for a sample of this see Figure \ref{fig:data_samples}) pose a challenge to the system's capability. These variations make robust navigation difficult, since a deployed robot will encounter circumstances it has never seen during training. A possible solution to this problem is to create a system, which is capable of adapting to new situations, while reusing as much of the existing knowledge as possible, and ideally only needing minimal information about the new domain to adapt to it.
    \par
    We framed this problem as a meta-learning problem, where we learn a base-learner, which is capable of few-shot adapting to a new environment. For this we used Model-Agnostic Meta-Learning (MAML) \citep{Finn2017}, particularly its variations MAML++ \citep{Antoniou2018} and ANIL \citep{Raghu2019}. We first establish that a model trained with MAML is capable of learning the underlying representations as well as a conventional approach. Then, we evaluate its adaptation capabilities when trained and validated only on a part of the data.
    \par
    We claim that: (i) Our MAML architecture is capable of learning the keypoints as well as the existing non-MAML system. (ii) Our MAML system is superior in adapting to unseen conditions, which it is capable of even in situations where the training domain is small and the domain shift big.


\section{Approach}
\label{sec:citations}

    \subsection{Dataset}
    We employed the dataset used in \citet{Sivakumar-RSS-24}, which consists of 28273 images. The images where taken on 54 different days, each attributed to either the early, late or very late season. Figure \ref{fig:data_samples} shows the spectrum of images present in the dataset. 
    \par 
    Each of the images has a corresponding label consisting of the coordinates of the three keypoints: vanishing point, left intersect-point with the x-axis and right intersect-point with the x-axis. 

    \begin{table}[h]
    \makebox[\textwidth]{
    \renewcommand{\arraystretch}{1.2}
    \begin{tabular}{l@{\extracolsep{20pt}}c@{\extracolsep{5pt}}c@{\extracolsep{20pt}}c@{\extracolsep{5pt}}c@{\extracolsep{20pt}}c@{\extracolsep{5pt}}c@{\extracolsep{0pt}}}
    \toprule
    & \multicolumn{2}{c}{Early Season} & \multicolumn{2}{c}{Late Season} & \multicolumn{2}{c}{Very Late Season} \\
    \cmidrule(){2-3} \cmidrule(){4-5} \cmidrule(){6-7}
    \multicolumn{1}{c}{\textbf{Train Split Name}} & \textbf{Days} & \textbf{Images} & \textbf{Days} & \textbf{Images} & \textbf{Days} & \textbf{Images} \\
    \midrule
    All-Season         & 13 & 6089 & 29 & 14897 & 1 & 2351 \\
    Early              & 13 & 6089 & 0  & 0     & 0 & 0    \\
    \bottomrule
    \end{tabular}
    }
    \vspace{\abovecaptionskip}
    \caption{The structure of the whole training dataset and the \textit{Early} subset.}
    \label{table:datasplits}
    \end{table}
    
    \subsection{Model}
    Our model uses a ResNet-18-based encoder \citep{Kaiming2015}, pre-trained on ImageNet \citep{Krizhevsky2012}, which together with a bilinearly upsampling decoder forms a U-Net-like \citep{Ronneberger2015} architecture.
    \par
    See \citet{Sivakumar-RSS-24} for more details regarding the model.

    \subsection{Training without MAML}
    The training without MAML is straightforward. For a given training split (see Table \ref{table:datasplits}), a dataloader is created where each of the images inside the days entailed by the split are equally likely to be sampled for a certain batch.
    \par
    We train with a learning rate of 1e-4 for $50$ epochs. 
    Apart from the fact that no data augmentations were used, our training process in the non-MAML case resembles that described by \citet{Sivakumar-RSS-24}.

    \subsection{Training with MAML}
    MAML requires the definition of tasks \citep{Finn2017}, from which a support set for inner optimization and a query set for outer loss calculation can be sampled. In our case, the tasks used in a given split are the different days it contains. To obtain a task from a training split, we sample a day from it, where the probability of a day being sampled is proportional to the number of images taken on said day, relative to all images in the split. After we have obtained the task, the support set and query set is created by sampling images taken during the day associated with this task.
    \par 
    We first tried to use vanilla MAML \citep{Finn2017}. However, it quickly emerged that vanilla MAML is not capable of learning the representations reliably, even in a setting where all available data is provided. The problem of MAML's sensitivity is discussed in \citet{Antoniou2018}, which also proposes concrete changes to the model training. In essence, these changes look at the inner optimization steps of MAML individually, learning separate learning rates and buffers for them among other optimizations (see Appendix \ref{appendix:MAML++-Improvements} for more details). Applying its techniques lead to a stabilized training and a resulting model, which performs better.
    \par 
    We also evaluate ANIL (Almost no Inner Loop) which was introduced in \citet{Raghu2019}. It proposes only updating the last layer inside the inner loop, making the finetuning process less computationally expensive while at times leading to better performance.


\begin{figure}[t]
    \centering
    \includegraphics[width=\textwidth]{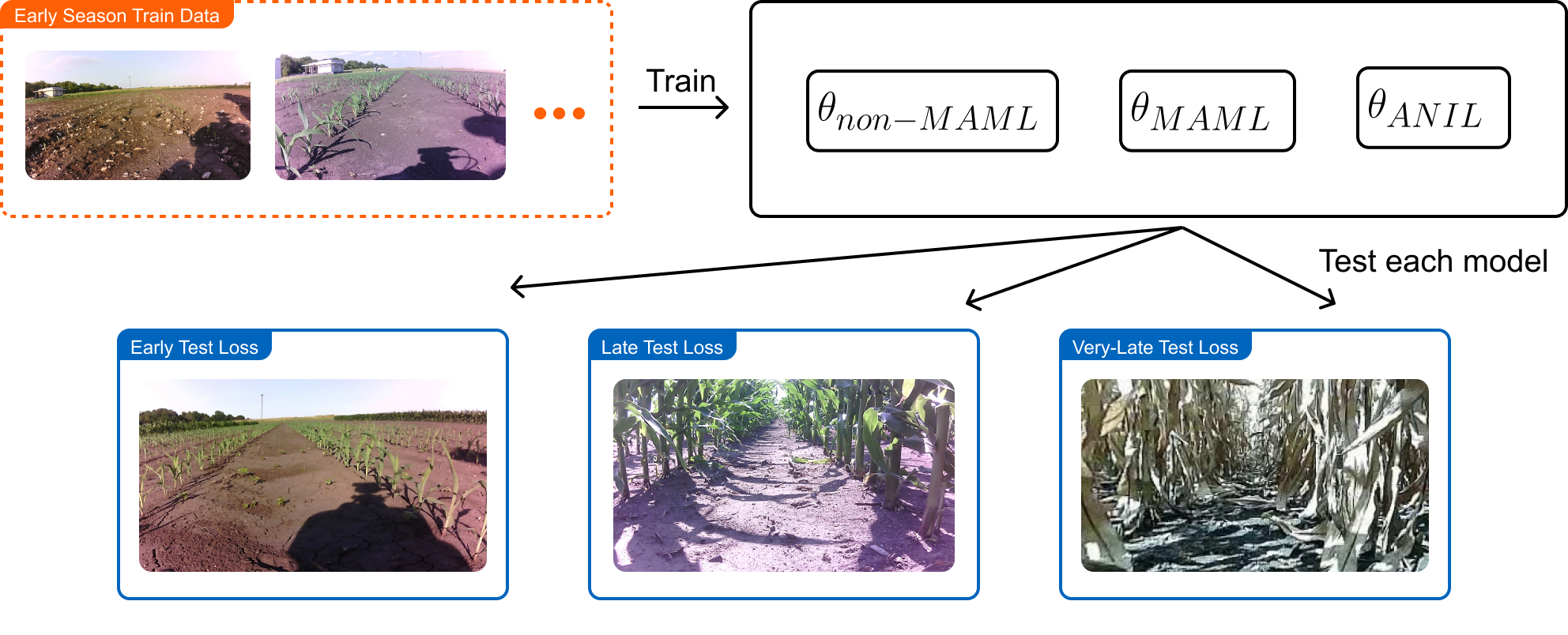}
    \caption{Our process starts by taking a subset of the data to train three models: Non-MAML, MAML, ANIL. For each of those, the validation set is from the same domain as the training set. We then see how they perform in the three domains (early, late and very late) calculating a test loss for each.} 
    \label{fig:summary}
\end{figure}

\section{Experimental Evaluation}
\label{sec:result}

    \subsection{Experiment Setup}
    \label{sec:Experiment-Setup}
    Our experiments are based on the different data splits presented in Table \ref{table:datasplits}. For each split, we do the following:
    \begin{enumerate}
        \item We run a MAML, ANIL and non-MAML training.
        \item For all of those trainings we choose the checkpoint which performs best in the domain the model was trained on. For example, for split \textit{Early}, we only look at the validation loss on the early val data to make a decision on which checkpoint is considered the best. 
        \par
        This is an important distinction compared to choosing the checkpoint which performs best across all validation data. Intuitively, it simulates a situation where we only have access to data from one season and have to make a decision about which model is best for other seasons without having data about them yet.
        \item We evaluate the chosen checkpoint on early, late and very late test data, resulting in 3 different losses. 
        \par
        In the case of non-MAML, we evaluated the model as is, as well as versions which were few-shot finetuned at different learning rates, using the same value of $k$ as we do in MAML, thus providing a fair assessment. In our case $k$ was $5$, meaning that all models got to use 5 images to finetune to a task at hand. 
        \par 
        In the case of MAML (or MAML++ to be precise), the output of the training process are the model weights of a base-learner, as well as the learning rates and buffers to be used during finetuning (see LSLR and BNRS in Appendix \ref{appendix:MAML++-Improvements}). The base-learner is then finetuned on $k = 5$ samples from the day it is being evaluated on. 
    \end{enumerate}

    The standard deviation values indicated in Table \ref{table:results} come from the element of randomness introduced by the fact that the models are finetuned at test-time and depending on the $k$ images used to finetune, the performance may vary. The standard deviation values provided are the result of three test runs respectively.

    \begin{table}[t]
    \makebox[\textwidth]{
    \renewcommand{\arraystretch}{1.2}
    \begin{tabular}{llccc}
    \toprule 
    \textbf{Train Split} & \textbf{Model} & \textbf{Early Test Loss} & \textbf{Late Test Loss} & \textbf{Very-Late Test Loss}\\
    \midrule
    All-Season & Non-MAML w/o finetune & $5.7$ & $7.7$ & $22.0$ \\
    & Non-MAML @ lr=0.1  & $5.9\pm0.2$ & $8.6\pm0.1$ & $24.0\pm0.3$ \\
    & MAML++              & $\bm{3.9\pm0.1}$ & $\bm{6.8\pm0.1}$ & $\bm{12.3\pm0.1}$ \\
    & ANIL++              & $4.3\pm0.0$ & $7.1\pm0.0$ & $13.6\pm0.1$ \\
    \midrule
    Early & Non-MAML w/o finetune & $4.8$ & $100.9$ & $118.7$ \\
    & Non-MAML @ lr=0.1 & $4.9\pm0.0$ & $59.7\pm2.1$ & $101\pm2.0$ \\
    & MAML++              & $\bm{2.8\pm0.0}$ & $\bm{28.8\pm1.1}$ & $\bm{43.2\pm1.9}$ \\
    & ANIL++              & $2.9\pm0.0$ & $76.3\pm1.2$ & $93.6\pm0.0$ \\
    \bottomrule
    \end{tabular}
    }
    \vspace{\abovecaptionskip}
    \caption{MAML slightly outperforms non-MAML when trained on data from all season, and significantly outperforms non-MAML if only trained on early season data. The reported loss values are the sum of L1-losses for the three keypoints. For each split (see Table \ref{table:datasplits}), we evaluate the Non-MAML, MAML++ and ANIL++ models. In the case of Non-MAML we evaluate it without finetuning and with finetuning at a learning rate of $0.1$ which was what worked best, outperforming the non-finetuned model at times. For additional runs on different datasplits see Appendix \ref{appendix:Additional-Splits-and-Runs}.}
    \label{table:results}
    \end{table}
    \subsection{Results}
    
    \subsubsection{Training on All-Season Data}
    Firstly, we establish that our MAML-based solution is capable of performing equally well as a non-MAML approach, when trained on the entire dataset (see Table \ref{table:results}). This was expected, proofs however that MAML++ and ANIL++ are capable of learning the representations of the under-canopy navigation problem.

    \subsubsection{Training Only on Early-Season Data}
\begin{figure}[t]
    \centering
    \includegraphics[width=0.9\textwidth]{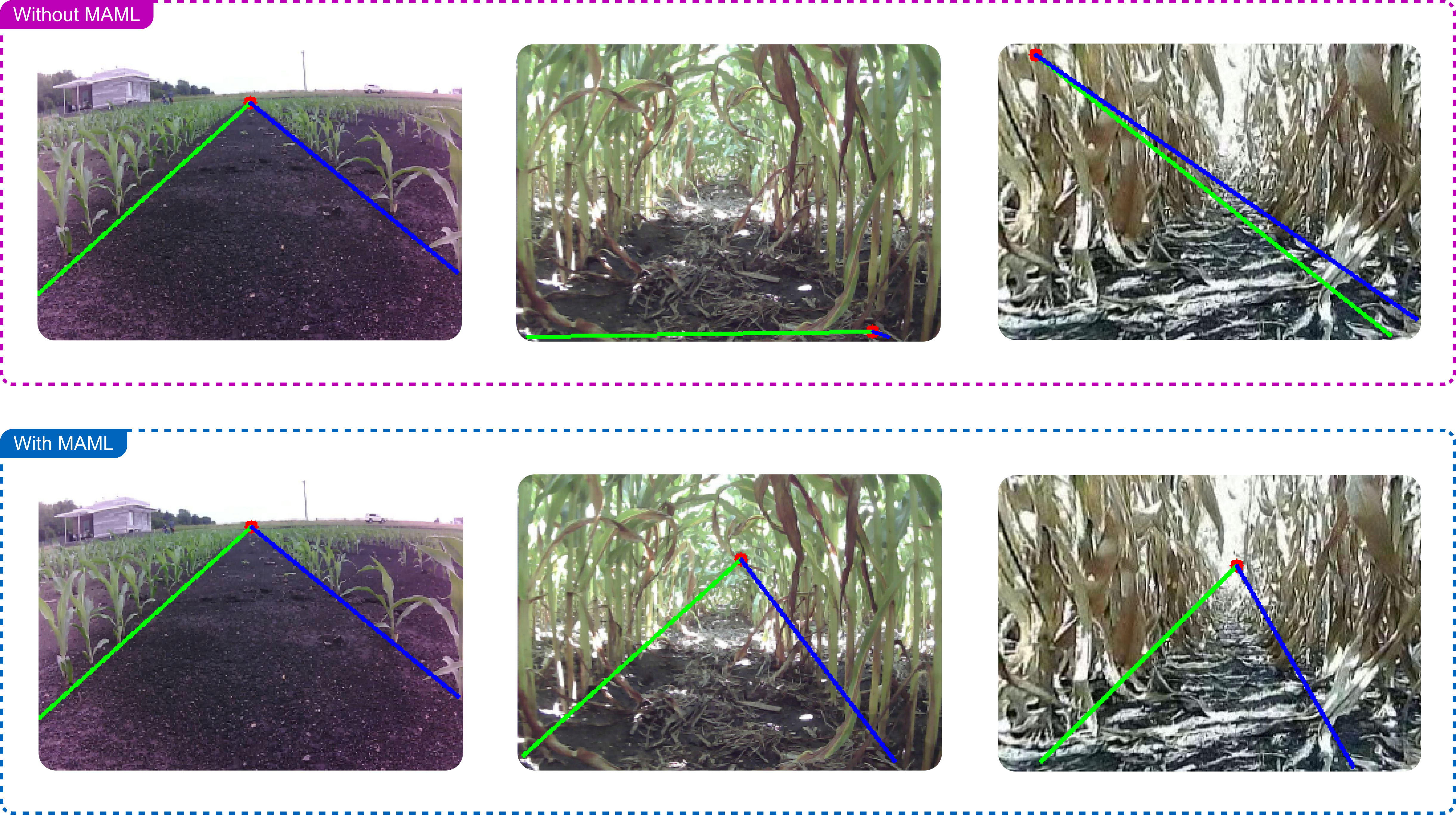}
    \caption{The test results of a model trained only on early-season data with MAML and without. Both models work good for early season, but only the MAML-trained model is capable of adapting to other seasons.} 
    \label{fig:maml-vs-nonmaml}
\end{figure}

    Furthermore, we show that our MAML++ system is capable of learning good representations for the whole season even if only trained on the data from one season (see Table \ref{table:results}). The non-MAML system struggles with this indicating that it is not capable of overcoming big domain shifts. See Figure \ref{fig:maml-vs-nonmaml} for a visual comparison of the two. 
    \par
    ANIL++ performs bad in this scenario, but there are checkpoints of the ANIL++ training on the \textit{Early} split, which perform almost as good as MAML++ in the late and very late season. However, those checkpoints perform marginally worse on the early data and as explained in \ref{sec:Experiment-Setup}, we always choose the checkpoint which performs best in the training domain, simulating a setting, where we only have data from this domain.

\section{Conclusion}
\label{sec:conclusion}

    In this paper, we explored the use of MAML in the case of under-canopy environments, showing that it is superior to a non-MAML approach in overcoming domain shifts using only minimal data from the target domain. 
    \par
    Nonetheless, a caveat of MAML remains the need for adaptation to each new task to reach its full performance, meaning a model does not work as-is after the training process. In the supervised case, this would mean that new images have to be labeled every time the robot enters a new environment. However, a inner loop learning in a self-supervised manner could solve this issue, leading to the robot adapting automatically to new environments \citep{wortsman2019}, which is why we see this as a promising direction for future work.


\clearpage
\acknowledgments 

This work was supported in part by USDA grant iFARM(\#2022-77038-37306) and utilizes resources supported by the National Science Foundation’s Major Research Instrumentation program, grant \#1725729, as well as the University of Illinois at Urbana-Champaign.


\bibliography{example}

\newpage
\appendix

\section{MAML++ Improvements}
    \label{appendix:MAML++-Improvements}
    The problem of MAML's sensitivity as we encountered it, is discussed in \citet{Antoniou2018}, which proposes the following changes:
    \begin{itemize}
        \item \textbf{Multi-Step Loss Optimization (MSL)} to fix the problem of gradient stability. It does so by creating the loss as a weighted sum of the different losses obtained after each inner update step instead of just the final step.
        \item \textbf{Derivative Order Annealing (DA)} to reduce computational cost. MAML requires the second order derivative, which is expensive to compute. MAML++ uses the first order approximation for a certain number of updates before switching to second order. The authors of the paper also found that this can lead to a more stabilized training. We made the same finding (see Appendix \ref{appendix:Hyperparameters}).
        \item \textbf{Per-Step Batch Normalization Running Statistics (BNRS)} to improve generalization performance and speed up optimization. Instead of a single set of running statistics, we accumulate $N$ sets (one for each of the $N$ optimization steps) during meta-training and use them for the respective steps during meta-test time.
        \item \textbf{Per-Step Batch Normalization Weights and Biases (BNWB)} to improve generalization performance and speed up optimization. Same reasoning and implementation as BNRS: a separate set of batch norm weights and biases for each of the $N$ inner optimization steps.
        \item \textbf{Learning Per-Layer Per-Step Learning Rates (LSLR)} to improve  generalization performance. Instead of a single inner learning rate $\alpha$, one learning rate for each step and each model layer is learned during meta-train time and - as above - used at meta-test time during finetuning.
        \item \textbf{Cosine-Annealing of Meta-Optimizer Learning Rate (CA)} to improve generalization performance. Instead of always using the same outer learning rate $\beta$, episode 1 starts out with $\beta$ and drops to $\beta_{min}$ by the last episode, following cosine annealing.
    \end{itemize}
    We implemented all of the above additions of MAML++, except for BNWB. The result was a model, which was better at generalizing and also more stable during training compared to vanilla MAML. 

\section{Hyperparameters}
    \label{appendix:Hyperparameters}
    Vanilla MAML already uses more hyperparameters than conventional training. MAML++ extends those with the percentage of iterations to do MSL and the percentage of iterations to do first order updates (DA).
    \par 
    We made the following experiences while trying to find the best set of hyperparameters:

    \begin{itemize}
        \item \textbf{Number of episodes:} All runs were trained with twenty thousand episodes. Oftentimes they would converge faster for the domain they were trained on (e.g. early season data) compared to other domains. This indicates, that even after the model has learned to perform well on a given task, it continues to learn underlying information, which helps when finetuning in new domains.
        \par
        The fact that we still trained with twenty thousand episodes and not less does not contradict the statement made in \ref{sec:Experiment-Setup} regarding choice of the best model based on the training domain performance only. While we made this interesting finding, we still stuck with our approach of choosing the model which is best in the training domain, independently of its performance in another domain.
        \item \textbf{Meta batch size:} The number of tasks per outer loop iteration. We consistently used $4$ for this throughout our training. 
        \item \textbf{Number of images to use in the inner loop ($k$):} We tried $k=1, 3, 5, 8$. And higher $k$ unsurprisingly lead to better domain adaptation, since more of the newer domain is learned. However, there is only a small difference between setting $k$ to one of $3, 5, 8$, whereas $k=1$ (i.e. only one picture to finetune on) performs notably worse. Here, a balance has to be struck between performance and a small number of images to be used during finetuning.
        \item \textbf{Number of inner steps ($N$):} We tried out $N = 1, 3, 5$, with all of them performing well, indicating a rather small difference between the different tasks. Interestingly, convergence takes longer the smaller the N is.
        \item \textbf{Inner learning rate ($\alpha$):} We used $0.4$ for this. However this is only an initialization in the cases where the LSLR addition from MAML++ is used, since those learning rates are gonna be learned along the training anyway. Still, $0.4$ seems to be the correct initialization for this, since runs with $\alpha = 0.1, 0.8$ performed worse.
        \item \textbf{Outer learning rate ($\beta$):} We used $0.001$ for this, and let it not drop below $0.00001$ when cosine annealing (CA) was used.
        \item \textbf{Percentage of episodes that use Multi-Step Loss (MSL)}: We found this value to have little effect on performance as long as it is high enough. The difference between $0.5$ and $0.99$ is small, however not having this at all leads to worse results and possibly instable runs.
        \item \textbf{Percentage of episodes that use first order updates (DA)}: For this hyperparameter the sweetspot seems to be around $0.3$ for our usecase. When it is too small, the training becomes unstable, and at the same time we made the experience that higher values lead to higher losses for domains with a big difference compared to the training domain (e.g. very late test vs early train data).
    \end{itemize}

    The final hyperparameters used for all runs are: Number of episodes: $20,000$, Meta batch size: $4$, $k$: $5$, $N$: 3, $\alpha$: $0.4$, $\beta$: $0.001$, Percentage of episodes using MSL: $0.99$, Percentage of episodes using first order updates: $0.3$. 

\section{Additional Splits and Runs}
    \label{appendix:Additional-Splits-and-Runs}
    Additionally to the \textit{All-Season} and \textit{Early} splits detailed in Table \ref{table:datasplits}, we also ran the comparison of MAML and non-MAML with different splits, namely ones involving less data in the respective training domains (see Table \ref{table:append_datasplits}). In all of the cases, we found MAML++ to perform better than non-MAML, especially in scenarios where we train with less data. 
    \par
    Notably, a MAML++ model trained with only $1372$ images from the late season is capable of roughly matching the performance of a non-MAML model trained with $23337$ images coming from all seasons.
    \par 
    See Table \ref{table:append_results} for details.
    
\begin{table}[t]
    \makebox[\textwidth]{
    \renewcommand{\arraystretch}{1.2}
    \begin{tabular}{l@{\extracolsep{20pt}}c@{\extracolsep{5pt}}c@{\extracolsep{20pt}}c@{\extracolsep{5pt}}c@{\extracolsep{20pt}}c@{\extracolsep{5pt}}c@{\extracolsep{0pt}}}
    \toprule
    & \multicolumn{2}{c}{Early Season} & \multicolumn{2}{c}{Late Season} & \multicolumn{2}{c}{Very Late Season} \\
    \cmidrule(){2-3} \cmidrule(){4-5} \cmidrule(){6-7}
    \multicolumn{1}{c}{\textbf{Train Split Name}} & \textbf{Days} & \textbf{Images} & \textbf{Days} & \textbf{Images} & \textbf{Days} & \textbf{Images} \\
    \midrule
    All-Season         & 13 & 6089 & 29 & 14897 & 1 & 2351 \\
    All-Season-Subset & 1  & 1636 & 1  & 1482  & 1 & 2351 \\
    \midrule
    Early              & 13 & 6089 & 0  & 0     & 0 & 0    \\
    Early-Subset1      & 1  & 1636 & 0  & 0     & 0 & 0    \\
    Early-Subset2      & 6  & 1544 & 0  & 0     & 0 & 0    \\
    \midrule
    Late               & 0 & 0 & 29  & 14897     & 0 & 0    \\
    Late-Subset       & 0  & 0 & 6  & 1372     & 0 & 0    \\
    \bottomrule
    \end{tabular}
    }
    \vspace{\abovecaptionskip}
    \caption{Additional datasplits, extending those in Table \ref{table:datasplits}.}
    \label{table:append_datasplits}
    \end{table}

\begin{table}[t]
    \makebox[\textwidth]{
    \renewcommand{\arraystretch}{1.2}
    \begin{tabular}{llccc}
    \toprule 
    \textbf{Train Split} & \textbf{Model} & \textbf{Early Test Loss} & \textbf{Late Test Loss} & \textbf{Very-Late Test Loss}\\
    \midrule 
    All-Season & Non-MAML w/o finetune & $5.7$ & $7.7$ & $22.0$ \\
    & Non-MAML @ lr=0.1  & $5.9\pm0.2$ & $8.6\pm0.1$ & $24.0\pm0.3$ \\
    & Non-MAML @ lr=0.4 & $107.4\pm26.3$ & $75.5\pm14.6$ & $78.1\pm8.9$ \\
    & MAML++              & $\bm{3.9\pm0.1}$ & $\bm{6.8\pm0.1}$ & $\bm{12.3\pm0.1}$ \\
    & ANIL++              & $4.3\pm0.0$ & $7.1\pm0.0$ & $13.6\pm0.1$ \\
    \midrule
    All-Season-Subset & Non-MAML w/o finetune & $24.2$ & $19.4$ & $27.1$ \\
    & Non-MAML @ lr=0.1 & $30.8\pm1.9$ & $21.6\pm0.3$ & $27.6\pm0.3$ \\
    & Non-MAML @ lr=0.4  & $179.0\pm12.2$ & $164.6\pm21.7$ & $183.6\pm37.9$ \\
    & MAML++              & $\bm{13.8\pm0.2}$ & $\bm{15.3\pm0.4}$ & $\bm{11.9\pm0.1}$ \\
    & ANIL++              & $17.7\pm0.3$ & $21.6\pm0.4$ & $12.2\pm0.1$ \\
    \midrule
    \midrule
    Early & Non-MAML w/o finetune & $4.8$ & $100.9$ & $118.7$ \\
    & Non-MAML @ lr=0.1 & $4.9\pm0.0$ & $59.7\pm2.1$ & $101\pm2.0$ \\
    & Non-MAML @ lr=0.4  & $169.2\pm33.4$ & $196.5\pm35.3$ & $232.4\pm16.2$ \\
    & MAML++              & $\bm{2.8\pm0.0}$ & $\bm{28.8\pm1.1}$ & $\bm{43.2\pm1.9}$ \\
    & ANIL++              & $2.9\pm0.0$ & $76.3\pm1.2$ & $93.6\pm0.0$ \\
    \midrule
    Early-Subset1 & Non-MAML w/o finetune & $27.6$ & $85.1$ & $103.0$ \\
    & Non-MAML @ lr=0.1 & $27.1\pm4.0$ & $69.0\pm0.4$ & $101.3\pm2.4$ \\
    & MAML++              & $\bm{9.6\pm0.4}$ & $\bm{34.3\pm2.5}$ & $\bm{73.9\pm4.5}$ \\
    & ANIL++              & $12.1\pm0.2$ & $76.5\pm0.4$ & $133.5\pm0.5$ \\
    \midrule
    Early-Subset2 & Non-MAML w/o finetune & $11.1$ & $86.2$ & $103.6$ \\
    & Non-MAML @ lr=0.1 & $8.2\pm0.4$ & $75.2\pm0.6$ & $99.3\pm4.9$ \\
    & MAML++              & $\bm{4.0\pm0.0}$ & $\bm{34.1\pm1.3}$ & $\bm{26.8\pm0.5}$ \\
    & ANIL++              & $4.1\pm0.0$ & $52.6\pm1.0$ & $87.4\pm1.6$ \\
    \midrule
    \midrule
    Late & Non-MAML w/o finetune & $16.9$ & $7.3$ & $48.3$ \\
    & Non-MAML @ lr=0.1 & $18.4\pm1.8$ & $11.9\pm0.4$ & $46.9\pm1.1$ \\
    & MAML++              & $\bm{9.7\pm0.3}$ & $6.4\pm0.0$ & $\bm{20.3\pm0.2}$ \\
    & ANIL++              & $11.2\pm0.0$ & $\bm{6.2\pm0.0}$ & $38.3\pm0.1$ \\
    \midrule
    Late-Subset & Non-MAML w/o finetune & $27.6$ & $31.9$ & $86.6$ \\
    & Non-MAML @ lr=0.1 & $30.5\pm2.2$ & $28.9\pm2.2$ & $80.6\pm2.2$ \\
    & MAML++              & $\bm{11.4\pm0.6}$ & $\bm{10.4\pm0.1}$ & $\bm{18.9\pm1.4}$ \\
    & ANIL++              & $12.5\pm0.0$ & $11.1\pm0.0$ & $36.9\pm0.3$ \\
    \bottomrule
    \end{tabular}
    }
    \caption{Additional runs on different datasplits (see Table \ref{table:append_datasplits}), following the same prodcedure as in Table \ref{table:results}. We tried finetuning the non-MAML model on learning $0.4$, since this is the initilization learning rate of MAML++ and ANIL++. This high learning rate leads to an entirely useless "finetuned" model.}
    \label{table:append_results}
    \end{table}

\end{document}